\documentclass[10pt,twocolumn,letterpaper]{article}

\usepackage{wacv}
\usepackage{times}
\usepackage{epsfig}
\usepackage{graphicx}
\usepackage{amsmath}
\usepackage{amssymb}
\usepackage{bm}
\usepackage{fixltx2e}
\usepackage{placeins}
\usepackage{gensymb}
\usepackage{tablefootnote}
\usepackage[multiple]{footmisc}
\usepackage[font=footnotesize]{caption}
\usepackage[pagebackref=true,breaklinks=true,letterpaper=true,colorlinks,bookmarks=false]{hyperref}
\usepackage[colorlinks]{hyperref}

\usepackage[pagebackref=true,breaklinks=true,letterpaper=true,colorlinks,bookmarks=false]{hyperref}

\wacvfinalcopy % *** Uncomment this line for the final submission

\def\wacvPaperID{****} % *** Enter the wacv Paper ID here

\ifwacvfinal\fi
\begin{document}

%%%%%%%%% TITLE
\title{On Hallucinating Context and Background Pixels from a Face Mask using Multi-scale GANs\thanks{* This work was done while SB was at Notre Dame}}

\author{\parbox{16cm}{\centering
    {\large Sandipan Banerjee*$^1$, Walter J. Scheirer$^2$, Kevin W. Bowyer$^2$, and Patrick J. Flynn$^2$}\\
    $^1$ Affectiva, USA\\
    $^2$ Department of Computer Science \& Engineering, University of Notre Dame, USA\\
        \tt\small sandipan.banerjee@affectiva.com,     \{wscheire, kwb, flynn\}@nd.edu
    }
    % <-this % stops a space
}

% \author{\parbox{16cm}{\centering
%     {\large Sandipan Banerjee*$^1$, Joel Brogan*$^1$, Janez Kri\v{z}aj$^2$, Aparna Bharati$^1$, Brandon RichardWebster$^1$, Vitomir \v{S}truc$^2$, Patrick J. Flynn$^1$ and Walter J. Scheirer$^1$}\\
%     {\normalsize
%     $^1$ Dept. of Computer Science \& Engineering, University of Notre Dame, USA\\
%     $^2$ Faculty of Electrical Engineering, University of Ljubljana, Slovenia\\
%         \tt\small \{sbanerj1, jbrogan4, abharati, brichar1, flynn, wscheire\}@nd.edu\\
%         \tt\small \{janez.krizaj, vitomir.struc\}@fe.uni-lj.si
%     }}
%     % <-this % stops a space
% }

\renewcommand\footnotemark{}
\renewcommand\footnoterule{}

% % Authors at the same institution
% %\author{First Author \hspace{2cm} Second Author \\
% %Institution1\\
% %{\tt\small firstauthor@i1.org}
% %}
% % Authors at different institutions
% \author{First Author \\
% \and
% Second Author \\

\maketitle
\ifwacvfinal\thispagestyle{empty}\fi

\begin{abstract}
We propose a multi-scale GAN model to hallucinate realistic context (forehead, hair, neck, clothes) and background pixels automatically from a single input face mask, without any user supervision. Instead of swapping a face on to an existing picture, our model directly generates realistic context and background pixels based on the features of the provided face mask. Unlike facial inpainting algorithms, it can generate realistic hallucinations even for a large number of missing pixels. Our model is composed of a cascaded network of GAN blocks, each tasked with hallucination of missing pixels at a particular resolution while guiding the synthesis process of the next GAN block. The hallucinated full face image is made photo-realistic by using a combination of reconstruction, perceptual, adversarial and identity preserving losses at each block of the network. With a set of extensive experiments, we demonstrate the effectiveness of our model in hallucinating context and background pixels from face masks varying in facial pose, expression and lighting, collected from multiple datasets subject disjoint with our training data. We also compare our method with popular face inpainting and face swapping models in terms of visual quality, realism and identity preservation. Additionally, we analyze our cascaded pipeline and compare it with the progressive growing of GANs, and explore its usage as a data augmentation module for training CNNs.

\end{abstract}
%%%%%%%%% BODY TEXT
\vspace{-0.3cm}
\section{Introduction}
% \cite{LFW,IJBB,VGG,MasiAug,masiFG17,GangHua,GAN,ProgressiveGAN,Ulyanov,ResNet,atrous,pixshuff,UNet,ZhangCVPR18,salimans,FID,ColorZhang,ColorTTIC,SREFIDonor,HeInit,pix2pix,Caffe,TF,chollet2015keras,PyTorch,DSS,zhou2016semantic,zhou2017scene,xiao2018unified,Adam,lsgan}
\begin{figure}[t]
\centering
   \includegraphics[width=1.0\linewidth]{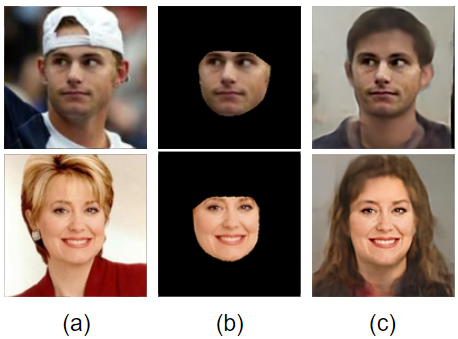}
   \caption{Our model, instead of swapping faces or inpainting missing facial pixels, directly hallucinates the entire context (forehead, hair, neck, clothes) and background from the input face mask. Sample results - (a) original face images from LFW \cite{LFW} (2D aligned), (b) corresponding face masks (input), and (c) the hallucinated output generated by our cascaded network of GANs trained on \cite{SREFIDonor}. All images are 128$\times$128 in size.}
\label{fig:Teaser1}
\vspace{-0.5cm}
\end{figure}
Generative adversarial nets (GANs) have revolutionized face synthesis research with algorithms being used to generate high quality synthetic face images \cite{DCGAN,BEGAN,ProgressiveGAN,StyleGen} or artificially edit visual attributes of existing face images like age \cite{cGAN,FaceAge}, pose \cite{DRGAN,LiuFrontal,TPGAN}, gender, expression and hairstyle \cite{GangHua,FSNet,AttnGAN}. However, these models require the full face image, comprising of the actual face, the context (forehead, hair, neck, clothes) and background pixels, to work. They fail to generate plausible results when the context and background pixels are absent (\ie, when only the face mask is present). Facial inpainting models \cite{GenFaceComp,CVPR17Yeh,GenFaceSIG17,HiResFaceComp,DeepFill,symmFC,EdgeConn} that inpaint `holes' work well when the missing pixels are small in number, located on or near the face. They do not generate realistic results when all of the context and background pixels are masked, as demonstrated in \cite{DFI} and the experiments in Section 4 of this paper. As a potential solution, we propose a cascaded GAN model that requires only a few thousand training face images to generate realistic synthetic context and background pixels from face masks with different gender, ethnicity, lighting, pose and expression, across different datasets. Our model can be used to generate - (1) supplemental training data for CNNs, adding variety to the hair and background for real subjects or synthetic face masks generated by \cite{MasiAug,SREFI2} (section 4.4 of this paper), and (2) stock images for media usage without any copyright and privacy concerns.

\begin{figure*}
\centering
   \includegraphics[width=1.0\linewidth]{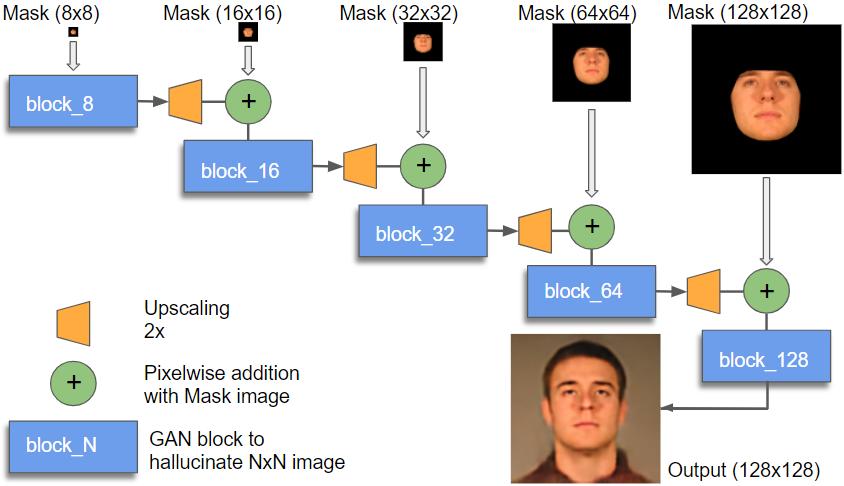}
   \caption{Our multi-scale cascaded network pipeline. Starting from the lowest resolution block (8$\times$8), we proceed higher up through a set of GAN blocks in a single pass (left to right in the figure). Except the last block, the output of each block is upscaled 2x and fed as input to the next block. To preserve fine facial details at each resolution, we add the mask image at each resolution before feeding the input. The final 128$\times$128 output, with hallucinated context and background pixels, is generated by block\_128. More details about the architecture of block\_128 is provided in Figure \ref{fig:Block128}.}
\label{fig:Pipeline}
\vspace{-0.5cm}
%\label{fig:onecol}
\end{figure*}

During training, our model takes as input a face image and its masked version, 128$\times$128 in size, and downsamples both to their 64$\times$64, 32$\times$32, 16$\times$16, and 8$\times$8 versions. Training starts at the lowest GAN block (block\_8), where it learns to reconstruct the 8$\times$8 full face image from the corresponding 8$\times$8 masked input. The output of this network is then upscaled 2x using a pixel shuffling block \cite{pixshuff} and passed to the next GAN block (block\_16). Thus instead of masked black pixels, block\_16 receives a 16$\times$16 input with roughly hallucinated context and background pixels, guiding it towards the direction of correct reconstruction. Its 16$\times$16 output is then upscaled and sent to block\_32 and so on (see Figure \ref{fig:Pipeline}). At each block, we independently learn to hallucinate context and background pixels through reconstruction loss, adversarial loss provided by a discriminator, perceptual loss from \cite{ZhangCVPR18} and an identity preserving loss using the pre-trained VGG-Face model \cite{VGG}. During testing we only use the trained generator and pixel shuffling blocks to hallucinate the final 128$\times$128 full face image from an input face mask. Sample results can be seen in Figure \ref{fig:Teaser1}.

% The background of this hallucinated face image can be further changed by segmenting out the foreground mask (the person) and blending in pixels from a different background image outside this mask.

% To measure the quality of our generated samples, we compute the distance between the Inception-v3 \cite{Inceptionv3} activation distributions for synthetic and real samples using the Fretchet Inception Distance (FID) \cite{FID}. A lower FID suggests that the synthetic samples look similar to the real face images. 

We perform the following experiments to assess the effectiveness of our model:

1. To gauge the effectiveness of our model in generating identity preserving, natural looking and diverse set of images we - (a) perform face matching experiments on \cite{LFW} using the ResNet-50 model \cite{ResNet}, (b) calculate SSIM \cite{SSIM} and perceptual error \cite{PieAPP} values, and (c) the FID \cite{FID} between original and hallucinated images.

2. Using the above metrics, we compare our model with popular facial inpainting algorithms - {\bf GenFace} \cite{GenFaceComp}, {\bf DeepFillv1} \cite{DeepFill}, {\bf SymmFCNet} \cite{symmFC}, and {\bf EdgeConnect} \cite{EdgeConn}.

3. We compare our model with the popular {\bf DeepFake}\footnote{\url{https://en.wikipedia.org/wiki/Deepfake}} face swapping application. Since it works only with tight face crops from a single identity, we train it on the LFW\cite{LFW} subject, \emph{George\_W\_Bush}, with the highest number of images (530). The trained network is used to synthesize source face crops, which are blended in the target face images.

4. We compare our single pass cascaded network with its progressively growing (ProGAN) version \cite{ProgressiveGAN}, where initial set of layers in the generator model are learned for a number of training epochs at the lowest resolution (8$\times$8), and then we add new layers to learn hallucination at a higher resolution (16$\times$16) and so on. 

5. Using the CASIA-WebFace \cite{CASIA} dataset, we evaluate the potential usage of our model as a data augmentation module for training CNNs.

The main contributions of our paper are as follows:

1. We propose a method that can automatically synthesize context and background pixels from a face mask, using a cascaded network of GAN blocks, without requiring any user annotation. Each block learns to hallucinate the masked pixels at multiple resolutions (8$\times$8 to 128$\times$128) via a weighted sum of reconstruction, adversarial, identity preserving and perceptual losses. Trained with a few thousand images, it can hallucinate full face images from different datasets with a wide variety in gender, ethnicity, facial pose, expression and lighting. 

2. We compare our model with recently proposed facial inpainting models \cite{GenFaceComp,DeepFill,symmFC,EdgeConn} and the DeepFake face swapping software. Our model generates photo-realistic results that produce higher quality scores (identity preservation, realism and visual quality) compared to the other algorithms on LFW \cite{LFW}.

3. We analyze the differences between the end-to-end training of our cascaded model with the ProGAN training regime from \cite{ProgressiveGAN} while keeping the network architecture, and factors like training data, hyper parameters, and loss function fixed. We show the cascaded architecture to benefit the hallucination process and generate sharper results.

4. We evaluate the potential application of our model as a generator of supplemental training data for CNNs, to augment the intra-class variance by adding diverse hair and backgrounds to existing subjects of the dataset. When trained on this augmented data, we show the ResNet-50 model \cite{ResNet} to produce a boost in test performance.

\section{Related Work}
{\bf Face synthesis}: While face synthesis research has greatly benefited from GANs \cite{GAN,DCGAN,BEGAN,ProgressiveGAN,StyleGen}, work in this domain began by simply combining neighborhood patches from different images to synthesize new faces \cite{ICME05,SREFI1}. Other methods include expression and attribute flow for synthesizing new views of a face \cite{SIGGRAPH09,SIGGRAPH11}. Many works have also explored the use of a 3D head model to generate synthetic views of a face or frontalize it to an uniform setting \cite{HassFront,MasiAug,Vito,SREFI2} while others have used GANs for this purpose \cite{TPGAN,DRGAN,LiuFrontal}. Researchers have also used deep learning models to reconstruct face images from their rough estimates \cite{Belanger,DFI,PixelNN} or with new attributes altogether \cite{GangHua,AttnGAN,StarGAN}. 

{\bf Face swapping}: The first face swapping pipeline was proposed in \cite{Bitouk08}, where a face is de-identified by blending together facial parts from other images. Many methods have modified this idea of recombining facial parts to generate synthetic images for de-identification or data augmentation \cite{ACCV14,SREFI1,IraPortrait}. In \cite{FaceSwap}, a 3D morphable model based shape estimation is used to segment the source face and fit it to the target image prior to blending. Instead of using facial textures, the method in \cite{FSNet}, uses latent variables from a deep network for face swapping. A style transfer \cite{Gatys} based face swapping approach was proposed in \cite{ICCV17Twitter}; but it requires the network to be trained on only one source subject at a time. DeepFake is another recent method for face swapping, where an autoencoder is trained to reconstruct tight face crops of a subject from its warped versions. This trained autoencoder is then used to hallucinate the source subject from different target face images. However, it works with one subject at a time and requires the target images to be highly constrained in visual attributes making it impractical for many real world applications.

{\bf Face inpainting}: Image inpainting started with \cite{Inpainting} transferring low-level features to small unknown regions from visible pixels. In \cite{BMVC04,InpaintingCVIU}, this idea is used to reconstruct facial parts in missing regions using a positive, local linear representation. A simple inpainting scheme was proposed in \cite{InpaintingCVIU}, which uses features like ethnicity, pose and expression to fill missing facial regions. GANs have also been used for image completion, \eg in \cite{GenFaceSIG17,GenFaceComp}, a generator is used to hallucinate masked pixels, with discriminators and parser networks refining the results. In \cite{CVPR17Yeh,DeepFill,EdgeConn}, information from the available data, surrounding image features, and edge structures are used for inpainting respectively. Facial symmetry is directly enforced in \cite{symmFC} to improve global consistency. In \cite{FaceShop,scFegan}, the inpainting process is guided by a rough sketch provided by the user. All these methods work well with small targeted masks\cite{OcularBMVC}, located on or near the face region, but perform poorly when a large masked area is presented\cite{DFI}, like the full context and background. 

When supplied with a face mask (\ie, limited data) the goal of our model is to automatically hallucinate realistic context and background pixels. While doing so the gender, ethnicity, pose, expression of the input subject should be preserved. While face swapping \cite{ICCV17Twitter,FaceSwap,FSNet} and face editing \cite{GangHua,AttnGAN} algorithms have dealt with transferring the face and facial attributes from one identity to another, they require - (1) the full face image to work, and (2) similarity in visual appearance, and pose for identity preservation. Unlike previous work, we treat this problem along the same lines as image colorization \cite{ColorZhang,ColorTTIC} and directly hallucinate the missing pixels taking cues from the input data without any involvement from the user.
\vspace{-0.3cm}
\section{Our Method}
%\vspace{-0.3cm}
Since there can be many plausible hallucinations from a single face mask, we control this unconstrained problem using the training data. When provided with a face mask $I^M$ during training, our model tunes its weights {\bf w} such that its generated output $G(I^M)$ looks similar to the original face image $I^{GT}$. The weights are parameterized by $I^{GT}$ itself and after a few training epochs, the model learns to generate $G(I^M)$ closely identical to $I^{GT}$. During testing, this trained model requires only a face mask ($I^M$), and not the full face image ($I^{GT}$), to hallucinate realistic context and background pixels from the learned representations.

\subsection{Network Architecture} 
\begin{figure}[t]
\centering
   \includegraphics[width=1.0\linewidth]{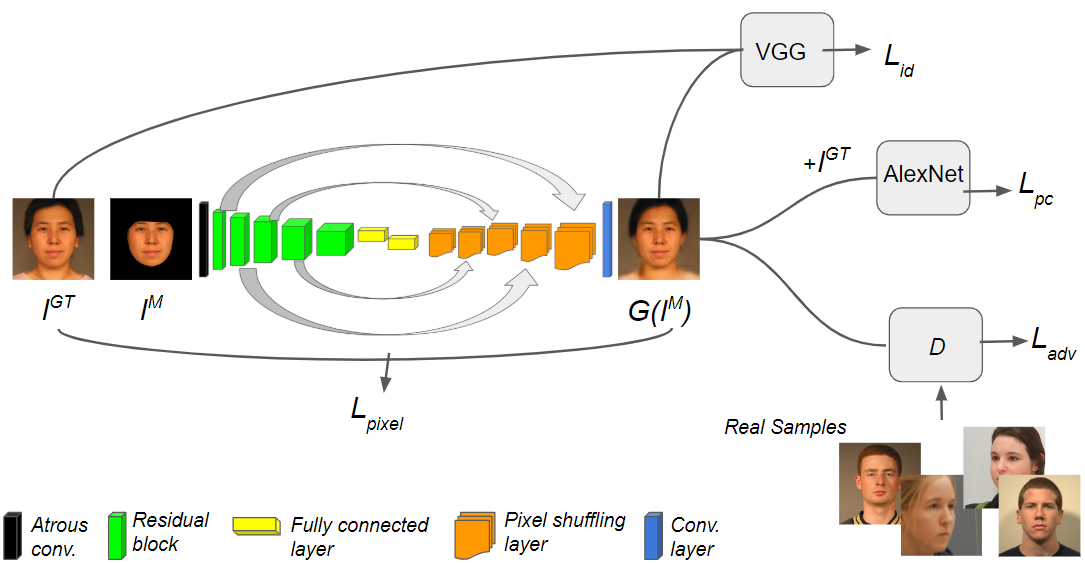}
   \caption{block\_128 architecture. The encoder is composed of five residual blocks while the decoder upsamples the encoded feature using five pixel shuffling blocks. The solid curved arrows between layers represent skip connections. During training the generator learns to hallucinate the original full face image $I^{GT}$ from the face mask $I^M$ via reconstruction, identity preserving, perceptual and adversarial losses. We replace pixels in the face mask of $G(I^{M})$ with original pixels from $I^{M}$ to preserve fine details.}
\label{fig:Block128}
\vspace{-0.5cm}
\end{figure}

{\bf Cascaded Network}. Inspired by \cite{Denton,Ulyanov,ICCV17Twitter}, we implement a multi-scale architecture comprising of five GAN blocks to learn hallucination at multiple resolutions (8$\times$8 to 128$\times$128), as depicted in Figure \ref{fig:Pipeline}. Unlike prior cascaded architectures, our model learns to hallucinate context and background for different image resolutions through a combination of multiple losses. Each block contains an encoder-decoder pair working as the generator. The encoder at the highest resolution block `block\_128', as shown in Figure \ref{fig:Block128}, takes the input and downsamples it through a set of strided convolution layers (stride = 2), except the first layer where we encapsulate extra spatial information using an atrous convolution layer \cite{atrous} with dilation rate of 2. Each of the next strided convolution layers is followed by a residual block \cite{ResNet} to facilitate the learning process. The output of the encoder is fed to the decoder which is composed of five convolution and pixel shuffling blocks \cite{pixshuff} for upscaling the feature by two in each dimension. 

We add skip connections \cite{UNet,ResNet,TPGAN} between encoder and decoder layers with the same tensor shape to propagate finer details from the input. The final 3 channel output is obtained by passing the upsampled result through a convolution layer with \emph{tanh} activation \cite{DCGAN,salimans}. Since the input and output of `block\_(N/2)' is half in height and width compared to `block\_N', each GAN block contains one fewer residual and pixel shuffling layers than its next GAN block. Except `block\_128', the output of each block is upscaled 2x through a pixel shuffling layer and fed as input to the next block. Thus, instead of a face mask, the block receives a rough hallucination to guide it towards the right direction. For all blocks, we also replace pixels in the face mask region of $G(I^M)$ with original pixels from $I^M$, before loss computation, to keep finer details of the face intact and focus only on the task of context and background generation.

During training, we provide each block with a discriminator to guide the generated samples towards the distribution of the training data. We use the popular \emph{CASIA-Net} architecture from \cite{CASIA} as the discriminator, after removing all max pooling and fully connected layers and adding batch normalization \cite{BatchNorm} to all convolution layers except the first one. A leaky \emph{ReLU} \cite{Relu} activation (slope = 0.2) is used for all layers except the last one where the \emph{sigmoid} activation is adopted to extract a probability between 0 (fake) and 1 (real), as suggested by \cite{DCGAN}. Each layer is initialized using He's initializer \cite{HeInit,ProgressiveGAN}. During testing, only the trained generator and pixel shuffling blocks are used to hallucinate the synthetic output, with resolution of 128$\times$128. 

\begin{figure}[t]
\centering
   \includegraphics[width=1.0\linewidth]{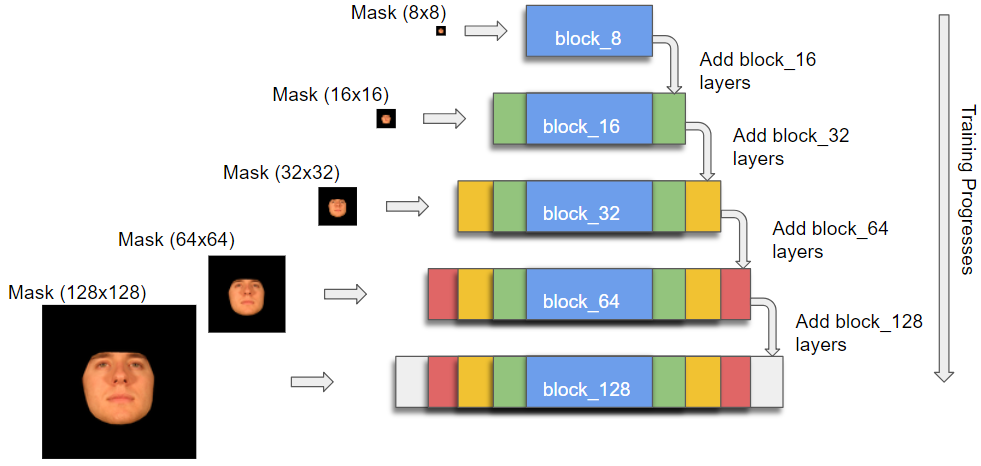}
   \caption{Pipeline of our progressively growing (ProGAN) network. We train the lowest resolution block for 50 epochs, then introduce additional layers for the next resolution block and resume training. This network growing continues till block\_128. During testing, we only use the trained block\_128.}
\label{fig:PGGAN}
\vspace{-0.5cm}
\end{figure}

{\bf Progressively Growing Network (ProGAN)}. Addressing the recently proposed progressive growing of GANs to generate high quality samples \cite{ProgressiveGAN,HiResFaceComp,StyleGen}, we also develop a ProGAN version of our model for comparison. Instead of the cascaded architecture where all the GAN blocks are trained in each iteration, we train the lowest resolution block\_8 first with 8$\times$8 face masks. After a few training epochs, we stop and load additional layers from block\_16 and start training again with 16$\times$16 face masks. This process of progressively growing the network by stopping and resuming training is continued till we have a trained block\_128 model, as depicted in Figure \ref{fig:PGGAN}. During testing, the trained block\_128 is used to hallucinate context and background pixels directly from previously unseen 128$\times$128 face masks. To maintain consistency, the loss function, hyper parameters and training data are kept the same with our cascaded network.

\subsection{Loss Function}
For each block of our network we learn context and background hallucinations independently. So we assign a combination of different losses, described below, to make the synthesized output at each resolution both realistic and identity preserving. We represent the image height, width and training batch size as $H$, $W$ and $N$ respectively.

1. {\bf Pixel loss ($L_{pixel}$)}: To enforce consistency between the pixels in the ground truth $I^{GT}$ and hallucinated face images $G(I^M)$, we adopt a mean $l_1$ loss computed as:
\begin{equation}
L_{pixel} = \frac{1}{N\times H\times W}\sum_{n=1}^{N}\sum_{i=1}^{H}\sum_{j=1}^{W} \left | (I^{GT}_{n})_{ij} - (G(I^{M}_{n}))_{ij} \right |
\label{eq:Lpix}
\end{equation}
where $H$ and $W$ increase as we move to higher blocks in our network, 8$\times$8 $\rightarrow$ 16$\times$16, 16$\times$16 $\rightarrow$ 32$\times$32, and so on. We use $l_1$ loss as it preserves high frequency signals better than $l_2$ in the normalized image thus generating sharper results.

2. {\bf Perceptual loss ($L_{pc}$)}: To make our hallucinations perceptually similar to real face images, we add the {\bf LPIPS} metric (ver. 0.0) from \cite{ZhangCVPR18} to our loss function. This metric finds a dissimilarity score between a pair of images, derived from deep features with varying levels of supervision, and is shown to be more consistent with human perception than classic similarity metrics like PSNR and SSIM \cite{SSIM}. We use LPIPS as a regularizer to support $L_{pixel}$. It is computed as:
\begin{equation}
L_{pc} = \frac{1}{N}\sum_{n=1}^{N}LPIPS(G(I^{M}_{n}), I^{GT}_{n})
\label{eq:Lpc}
\end{equation}
where $LPIPS$ is the dissimilarity score generated by the AlexNet \cite{AlexNet} model\footnote{Available here: \url{https://github.com/richzhang/PerceptualSimilarity}} (in PyTorch \cite{PyTorch}) provided by the authors. An $L_{pc}$ value of 0 suggests perfect similarity between $G(I^M)$ and $I^{GT}$. Since the code does not support low-res images, $L_{pc}$ is not applied on `block\_8' and `block\_16'.

3. {\bf Adversarial loss ($L_{adv}$)}: To push our hallucinations towards the manifold of real face images, we introduce an adversarial loss. This is achieved by training a discriminator along with the generator (encoder-decoder) at each block of our network. We use a mean square error based LSGAN \cite{lsgan} for this work as it has been shown to be more stable than binary cross entropy \cite{GAN}. The loss is calculated as:
\begin{equation}
L_{adv} = \frac{1}{N}\sum_{n=1}^{N}\left ( D(G(I^{M}_{n})) - c \right )^{2}
\label{eq:Ladv}
%\vspace{-0.2cm}
\end{equation}
where $D$ is the discriminator and $c$ is set to 1 as we want to fool $D$ into labeling the synthetic images as real.

4. {\bf Identity loss ($L_{id}$)}: To preserve essential features of the identity in the input face mask in the generated output, we use the pre-trained VGG-Face \cite{VGG} model to provide a supporting metric. We calculate the $l_2$ distance between the \emph{fc7} layer features between $I^{GT}$ and $G(I^M)$ and apply that as content loss similar to neural style transfer \cite{Gatys}. The closer this metric moves towards 0, the better the hallucination quality. The loss is calculated as:
\begin{equation}
L_{id} = \frac{1}{N\times \#F}\sum_{n=1}^{N}\sum_{i=1}^{\#F}(F(G(I^M_n))_i - F(I^{GT}_n)_i)^2
\label{eq:Lid}
%\vspace{-0.3cm}
\end{equation}
where $F$ is the 4096-D feature vector from VGG-Face \cite{VGG}. 

5. {\bf Total variation loss ($L_{tv}$)}: Similar to \cite{Johnson,TPGAN,ICCV17Twitter}, we add a total variation loss as a regularizer to suppress spike artifacts, calculated as:
\begin{multline}
L_{tv} = \sum_{i=i}^{H}\sum_{j=1}^{W}(G(I^M)_{i,j+1}-G(I^M)_{i,j})^2 + \\ (G(I^M)_{i+1,j}-G(I^M)_{i,j})^2 
\end{multline}
\label{eq:Ltv}
The final loss $L$ is computed as the weighted sum of the different losses:
\begin{equation}
L = L_{pixel} + \lambda_{1}L_{pc} + \lambda_{2}L_{adv} + \lambda_{3}L_{id} + \lambda_{4}L_{tv}
\label{eq:Lfull}
\vspace{-0.1cm}
\end{equation}

\section{Experiments}
{\bf Training Data.}For training our model, we randomly sample 12,622 face images (7,761 male and 4,861 female) from the public dataset in \cite{SREFIDonor}. These images were acquired specifically for recognition tasks, with variety of facial pose and neutral background. Image mirroring is then applied for data augmentation. To acquire the face masks, we first detect the face region using Dlib \cite{Dlib} and estimate its 68 facial keypoints with the pre-trained model from \cite{BulatLandmark}. We remove images that Dlib fails to detect a face from. The eye centers are then used to align the faces and pixels outside the convex hull of the facial landmark points in the aligned image is masked. Both the aligned and masked versions are then resized using bilinear interpolation to 8$\times$8$\times$3, 16$\times$16$\times$3, 32$\times$32$\times$3, 64$\times$64$\times$3 and 128$\times$128$\times$3, with pixels normalized between [0,1], for training different network blocks.

% \begin{figure*}
% \centering
% %\fbox{\rule{0pt}{2in} \rule{0.9\linewidth}{0pt}}
%   \includegraphics[width=1.0\linewidth]{Images/IJBB_Res2.png}
%   \caption{Sample results from IJB-B \cite{IJBB} (128$\times$128 in size). Top row - input face masks, Bottom row - full face images synthesized by our cascaded model.}
% \label{fig:IJBB_Res}
% %\label{fig:onecol}
% \vspace{-0.5cm}
% \end{figure*}

\begin{figure*}
\centering
  \includegraphics[width=1.0\linewidth]{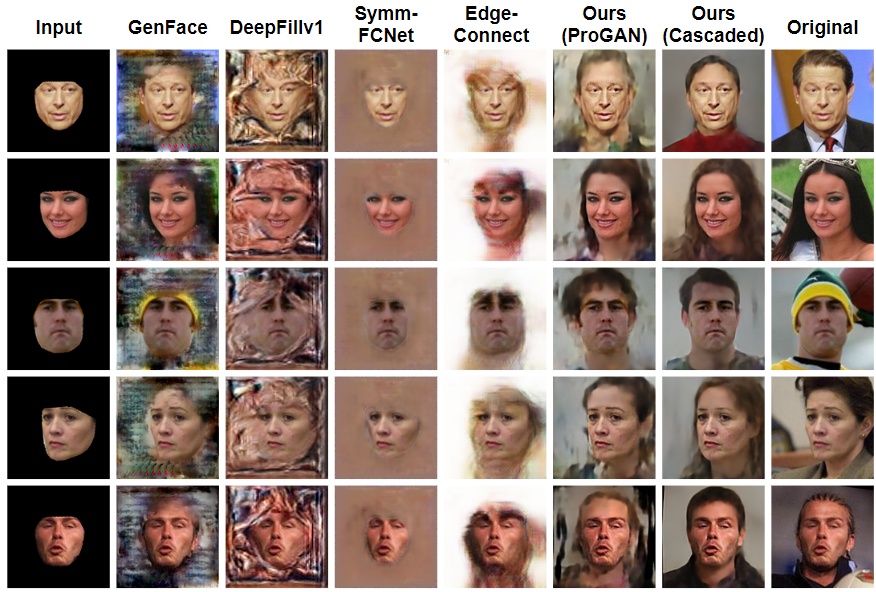}
  \caption{Sample results from LFW \cite{LFW} (128$\times$128 in size), generated using GenFace \cite{GenFaceComp}, DeepFillv1 \cite{DeepFill}, SymmFCNet \cite{symmFC}, EdgeConnect \cite{EdgeConn}, and our cascaded and ProGAN \cite{ProgressiveGAN} models. Note the variation in gender, pose, age, expression and lighting in the input images.}
\label{fig:LFW_BigImg}
%\label{fig:onecol}
\vspace{-0.5cm}
\end{figure*}

{\bf Training Details.} We train our model with the Adam optimizer \cite{Adam} with generator and discriminator learning rates set as $10^{-4}$ and $2\times 10^{-4}$ respectively. For each block, we train its discriminator with separate real and synthesized mini-batches with label smoothing applied to the real mini-batch, as suggested by \cite{DCGAN,salimans}. Other hyper-parameters are set empirically as $\lambda_1$ = 1, $\lambda_2$ = 0.1, $\lambda_3$ = 10, $\lambda_4$ = $10^{-6}$. We train our model on the NVIDIA Titan Xp GPU, using Tensorflow \cite{TF} and Keras \cite{chollet2015keras}, with a batch size of 10, for a hard limit of 50 epochs, as we find validation loss to plateau around this stage. We use the trained generator and pixel shuffling blocks from this model for our experiments.

{\bf Metrics for Quality Estimation.} To evaluate the effectiveness of our model in the task of context and background hallucination, and compare with other works, we use the following metrics:\\
(1) {\bf Mean Match Score}: We use the 256-dimensional penultimate layer descriptor from the `ResNet-50-256D' model \cite{ResNet}(`ResNet-50' here on), pre-trained on VGGFace2\cite{VGGFace2}\footnote{Available here: \url{https://github.com/ox-vgg/vgg_face2}}, as feature representation for an image for all our face recognition experiments. The deep features are extracted for each original image and the hallucinated output in the dataset. The mean match score $\rho$ is calculated by averaging the Pearson correlation coefficient between each feature pair as:
\begin{equation}
\rho = \frac{1}{N}\sum_{i=1}^{N}\frac{Cov((F_{o})_{i},(F_{h})_{i})}{\sigma_{(F_{o})_{i}}\sigma_{(F_{h})_{i}}}\label{eq:MeanMatch}
%\vspace{-0.1cm}
\end{equation}
where \emph{Cov} denotes covariance, \emph{N} is the number of images in the dataset, and $(F_{o})_{i}$ and $(F_{h})_{i}$ are the feature vectors of the i-th original and hallucinated images respectively. Ideally, we would like the hallucinated images to match well, but not perfectly, with the original images \ie, $\rho$ should be a little less than 1. Such a value would suggest that our model retains vital facial features of the input identity while adding variations in its visual attributes. The more the source face is modified, the more the gap widens, as specified in \cite{FaceSwap}.

\begin{table*}
\begin{center}
\captionsetup{justification=centering}
\caption{Quantitative results on the LFW \cite{LFW} dataset.}
\begin{small}
\begin{tabular}{  | c | c| c| c| c| }
\hline
\begin{tabular}[x]{@{}c@{}}{\bf Model}\end{tabular} & \begin{tabular}[x]{@{}c@{}}{\bf Mean Match Score}\end{tabular} & \begin{tabular}[x]{@{}c@{}}{\bf Mean SSIM} \cite{SSIM}\end{tabular} & \begin{tabular}[x]{@{}c@{}}{\bf FID} \cite{FID}\end{tabular} & \begin{tabular}[x]{@{}c@{}}{\bf Mean Perceptual Error} \cite{PieAPP}\end{tabular}\\
 %Ethnicity & Male images & Female images \\
\hline
\hline
  \begin{tabular}[x]{@{}c@{}}GenFace \cite{GenFaceComp}\end{tabular}  & 0.543 & 0.491 & 177.06 & 3.536 \\
  \hline
    \begin{tabular}[x]{@{}c@{}}DeepFillv1 \cite{CVPR17Yeh}\end{tabular}  & 0.481 & 0.321 & 241.696 & 3.204 \\
    \hline
    \begin{tabular}[x]{@{}c@{}}SymmFCNet \cite{symmFC}\end{tabular}  & 0.457 & 0.333 & 207.117 & 2.434 \\
    \hline
    \begin{tabular}[x]{@{}c@{}}EdgeConnect \cite{EdgeConn}\end{tabular}  & 0.454 & 0.178 & 141.695 & 3.106 \\
  \hline
  \hline
  \begin{tabular}[x]{@{}c@{}}DeepFake\end{tabular}  & 0.459 & 0.448 & {\bf 43.03} & 1.857 \\
  \hline
  \hline
    \begin{tabular}[x]{@{}c@{}}Ours (ProGAN)\end{tabular}  & 0.668 & 0.466 & 103.71 & 2.255 \\
    \hline
      \begin{tabular}[x]{@{}c@{}}Ours (Cascaded)\end{tabular}  & {\bf 0.722} & {\bf 0.753} & 46.12 & {\bf 1.256} \\
  \hline
\end{tabular}
\label{Tab:LFW_BigTab}
\end{small}
\end{center}
\vspace{-0.5cm}
\end{table*}

(2) {\bf Mean SSIM}: To evaluate the degree of degradation, or noise, in the hallucinated output, we compute the SSIM \cite{SSIM} value for each (original,synthetic) image pair in the dataset. A higher mean SSIM value suggests less noisy hallucinations and therefore a better model.

(3) {\bf FID}: To evaluate the realism of the generated samples, we use the Frechet Inception Distance (FID) metric proposed in \cite{FID}. FID uses activations from the Inception-v3 \cite{Inceptionv3} network to compare the statistics of the generated dataset to the real one. A lower FID suggests generated samples to be more realistic, and signifies a better model.

(4) {\bf Mean Perceptual Error}: To evaluate the perceptual dissimilarity between the original and the hallucinated images, we use the {\bf PieAPP} v0.1 metric using the pre-trained model from \cite{PieAPP}. The metric calculates the level of distortion between a pair of images, using a network trained on human ratings. A lower mean perceptual error indicates less noise in the hallucinated output, therefore a better model.

\subsection{Comparison with Facial Inpainting Models} 
To gauge how our model compares with algorithms for generating missing pixels, we make use of four popular facial inpainting models: {\bf GenFace} \cite{GenFaceComp}, {\bf DeepFillv1} \cite{DeepFill}, {\bf SymmFCNet} \cite{symmFC}, and {\bf EdgeConnect} \cite{EdgeConn}. We choose these models for our experiments, as - (1) they are open source with a pre-trained (on face images from CelebA \cite{celebA}) models available for use, unlike \cite{GenFaceSIG17,HiResFaceComp,FaceShop}, (2) they can work with 128$\times$128 face images, unlike \cite{CVPR17Yeh}, and (3) require no any user annotation, unlike \cite{scFegan}. 

To compare the models, we generate hallucinations using face masks from LFW \cite{LFW}. Since each model is trained with different binary masks of missing pixels, we provide the model a binary mask with every pixel outside the face labeled as `0' instead of the actual masked face we feed to our trained model. Both qualitative and quantitative comparisons can be seen in Fig. \ref{fig:LFW_BigImg} and Table \ref{Tab:LFW_BigTab} respectively. As shown in the table, our model (both versions) performs much better than the inpainting models for all metrics. These models aim to hallucinate the missing pixels, usually on or near the face region, using visual cues provided by facial pixels available in the image. Such cues are absent when whole of the context and background is masked, leading to noisy output. On the other hand, our model is specifically trained, and better suited for this task.
% Clearly, these models cannot handle large areas outside the face region and generate noise instead. It aims to hallucinate the missing pixels, usually on or near the face region, using visual cues provided by facial pixels available in the image. Such cues are absent when all the context and background pixels are masked, instead of small facial parts. Our model is specifically trained, and therefore better suited, for this task. Our goal is not to demean any existing algorithm but to show the effectiveness of our method in its task. We feel drawing analogy by comparing with both \emph{Deep Fakes}, which does swap the whole facial context and background, and inpainting algorithms provides a complete picture to the reader. 

\subsection{Comparison with DeepFake Face Swap} 
\begin{figure}[t]
\centering
   \includegraphics[width=1.0\linewidth]{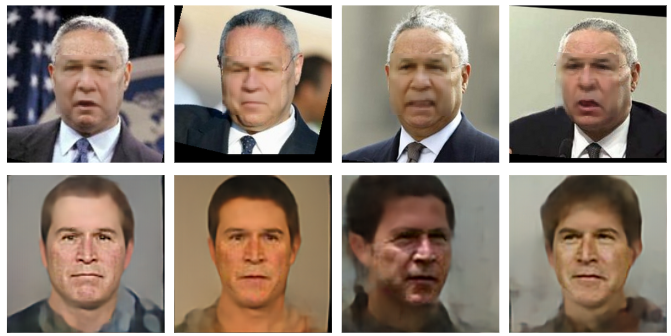}
   \caption{Top row - synthetic images generated using {\bf DeepFake} where the face mask (rectangle) is from `George\_W\_Bush' but the context and background are from real face images of `Colin\_Powell' (from LFW \cite{LFW}). Bottom row - synthesized context and background, using our trained cascaded model, for some images of the subject `George\_W\_Bush'.}
\label{fig:DF_Comp}
\vspace{-0.5cm}
\end{figure}

Owing to its huge popularity, we compare our model against the {\bf DeepFake} face swapping application. The software essentially trains an autoencoder to learn transformations to change an input face crop (target) to another identity (source) while keeping target visual attributes intact. Since this autoencoder learns transformations for one subject at a time, we train it using 64$\times$64 tight face crops of `George\_W\_Bush', the LFW\cite{LFW} identity with the most images (530). The autoencoder\footnote{We use the implementation from the most popular repo: \url{https://github.com/deepfakes/faceswap}} is trained for 10K iterations using these 530 images, following which it can be used to hallucinate images of `George\_W\_Bush' from face crops of other subjects and then blended onto the target images. The results of such a face swapping process can be seen in Figure \ref{fig:DF_Comp} where we swap `George\_W\_Bush' face images onto the context and background of `Colin\_Powell'. We choose `Colin\_Powell' as the mean hypercolumn \cite{HypCol} descriptor of his images, using \emph{conv}-[$1_2$,$2_2$,$3_3$,$4_3$,$5_3$] features from VGG-Face \cite{VGG}, is proximal to that of `George\_W\_Bush'. 

Although DeepFake produces plausible results (lower FID \cite{FID} in Table \ref{Tab:LFW_BigTab}), it requires both the source and target subjects to have fairly similar skin tone, pose and expression. Without such tight constraints, artifacts at the boundary of the blending mask are present as can be seen in the top row of Figure \ref{fig:DF_Comp} due to the difference in skin tone and absence of eyeglasses in the source identity. Our model, on the other hand, has no such constraints as it learns to hallucinate the full set of context and background pixels from the provided face mask itself. Also, our model achieves a higher mean match score than DeepFake suggesting that it preserves more discriminative features of the source in the hallucinated images while adding variations in appearance.

\subsection{Comparison with our ProGAN Model} 
For the progressively growing (ProGAN \cite{ProgressiveGAN}) version of our model, we set a training interval of 50 epochs after which we add new layers to the current block and resume training. Compared to the 96.53 hours required to train our cascaded network, our ProGAN model requires 66.24 hours to complete the full training at all scales, when trained on the same Titan Xp GPU system. The absence of multi-scale training, upscaling between blocks and depth concatenations during each iteration is the reason behind its lower training time. At the end of training, we feed 128$\times$128 face masks to block\_128 and get the hallucinated face images at the same resolution. We compare our cascaded and ProGAN models using masked face images from LFW \cite{LFW}; the quantitative results are shown in Table \ref{Tab:LFW_BigTab} and few qualitative samples can be seen in Figure \ref{fig:LFW_BigImg}.

Although the ProGAN model hallucinates slightly sharper results than the cascaded model due to the absence of upscaling between GAN blocks, it suffers from blurry artifacts, especially in the hair. This can be attributed to the fact that we only use block\_128 of the ProGAN model to synthesize the output directly at of 128$\times$128 like the trained generator from a single resolution GAN. Since the hallucination process in the cascaded network is guided at each resolution by the previous block, such artifacts are less frequent in its case. This might also be the reason of the difference in FID and perceptual error values between the two models in Table \ref{Tab:LFW_BigTab}.

\subsection{Effectiveness as Supplemental Training Data}
To evaluate if our model can be used to augment existing face image datasets, we perform a recognition experiment using the CASIA-WebFace (CW) dataset \cite{CASIA}. CW contains 494,414 face images of 10,575 real identities collected from the web. We align, mask and resize all the face images from CW using the same pre-processing steps as our training data. These masked images are then fed to our trained cascaded model to hallucinate synthetic context and background pixels. Since the identity of the input face mask is preserved in our model (as shown by the Mean Match Score in Table \ref{Tab:LFW_BigTab}), we label the hallucinated image as the same class as the original input from CW, similar to \cite{MasiAug,masiFG17,SREFI2}. In this way, we generate 494,414 synthetic images, with hallucinated context and background, from 494,414 existing images of 10,575 real identities. We prepare two training sets from the images - 1) a dataset containing 494,414 real images from CW and no synthetic images (Dataset 1 from Table \ref{Tab:LFW_CW}), and 2) a dataset containing 494,414 real images and 494,414 synthetic images of the same 10,575 subjects (Dataset 2 from Table \ref{Tab:LFW_CW}). 

\begin{table}
\begin{center}
\captionsetup{justification=centering}
\caption{Distribution and performance of training datasets with and without augmentation using our model.}
\begin{small}
\begin{tabular}{  | c | c| c| c| }
\hline
\begin{tabular}[x]{@{}c@{}}{\bf Training}\\{\bf Data}\end{tabular} & \begin{tabular}[x]{@{}c@{}}{\bf CW \cite{CASIA}}\\{\bf Images}\\(Identities)\end{tabular} & \begin{tabular}[x]{@{}c@{}}{\bf Hallucinated}\\{\bf Images}\\(Identities)\end{tabular} & \begin{tabular}[x]{@{}c@{}}{\bf LFW \cite{LFW}}\\{\bf Performance}\\(TPR@FPR = 0.01)\end{tabular}\\
 %Ethnicity & Male images & Female images \\
\hline
\hline
  \begin{tabular}[x]{@{}c@{}}Dataset 1 \end{tabular}  & \begin{tabular}[x]{@{}c@{}}494,414\\(10,575)\end{tabular} & 0 & 0.963 \\
  \hline
    \begin{tabular}[x]{@{}c@{}}Dataset 2\end{tabular}  & \begin{tabular}[x]{@{}c@{}}494,414\\(10,575)\end{tabular} & \begin{tabular}[x]{@{}c@{}}494,414\\(10,575)\end{tabular} & 0.971 \\
  \hline
\end{tabular}
\label{Tab:LFW_CW}
\end{small}
\end{center}
\vspace{-0.9cm}
\end{table}

% \begin{table}
% \begin{center}
% \captionsetup{justification=centering}
% \caption{Distribution and performance of training datasets with and without augmentation using our model.}
% \begin{small}
% \begin{tabular}{  | c | c| c| c|  }
% \hline
% \begin{tabular}[x]{@{}c@{}}{\bf Training}\\{\bf Data}\end{tabular} & 
% \begin{tabular}[x]{@{}c@{}}{\bf CW}\cite{CASIA}\\{\bf Images}\\(identities)\end{tabular} & 
% \begin{tabular}[x]{@{}c@{}}{\bf Synthetic}\\{\bf images}\\(identities)\end{tabular} &
% \begin{tabular}[x]{@{}c@{}}{\bf LFW}\cite{LFW}\\{\bf Performance}\\(TPR@FPR = 0.01)\end{tabular} \\
%  %Ethnicity & Male images & Female images \\
% \hline
% \hline
%   Dataset 1  &  494,414\\(10,575)  &  0  &  0.963\\
%   \hline
%   Dataset 2  &  494,414\\(10,575)  &  494,414\\(10,575)  &  0.971\\
% %   \hline
% %   Total  &  9011 (778)  &  6796 (674) \\
% \hline
% \end{tabular}
% \label{Tab:CWExp}
% \end{small}
% \end{center}
% \end{table}

We fine-tune the ResNet-50 \cite{ResNet} model with these datasets in two separate training sessions, where 90\% of the data is used for training and the rest for validation. The networks are trained using the Caffe \cite{Caffe} framework, with a base learning rate = 0.001 and a polynomial decay policy where gamma = 0.96, momentum = 0.009, and step size = 32K training iterations. We set the batch size = 16, and train each network till its validation loss plateaus across an epoch. After training terminates, we save its snapshot for testing on the LFW dataset \cite{LFW}. Each image is passed to the snapshot and its 256-D vector is extracted from the penultimate (\emph{feat\_extract}) layer. We use these features to perform a verification experiment (all vs. all matching) with Pearson correlation for scoring, the results of which are presented in Table \ref{Tab:LFW_CW}. As shown, the supplemental synthetic images introduce more intra-subject variation in context and background, which in turn slightly boosts the performance of the network during testing. Our trained model can therefore be used to augment existing face image datasets for training CNNs, especially to generate the diverse context and background pixels in synthetic face masks generated by \cite{MasiAug,SREFI2}.

\section{Detailed Model Architecture}
In this section, we list the layers of each generator block of our model. For both the cascaded and progressively growing (ProGAN) \cite{ProgressiveGAN} versions of our model, the architectures of the generator block remain the same. For the cascaded model however, we use a set of four pixel shuffling \cite{pixshuff} blocks to upscale the hallucination of a block 2x before feeding it as input to the next generator block. The architecture of each upscaling pixel shuffling blocks remains the same. The detailed layers of `block\_8', `block\_16', `block\_32', `block\_64', and `block\_128' layers are listed in Tables \ref{Tab:block8}, \ref{Tab:block16}, \ref{Tab:block32}, \ref{Tab:block64}, and \ref{Tab:block128} respectively. The convolution layers, residual blocks and pixel shuffling layers are indicated as `conv', 'RB', and `PS' respectively in the tables. For each of these layers in the generator, we used leaky \emph{ReLU} with slope of 0.1 as the activation, except for the last `conv' layer where a \emph{tanh} activation is used \cite{DCGAN,salimans}.
\begin{table}
\begin{center}
\captionsetup{justification=centering}
\caption{block\_8 architecture (input size is 8$\times$8$\times$3)}
\begin{small}
\begin{tabular}{  | c | c| c| }
\hline
\begin{tabular}[x]{@{}c@{}}{\bf Layer}\end{tabular} & \begin{tabular}[x]{@{}c@{}}{\bf Filter/Stride/Dilation}\end{tabular} & \begin{tabular}[x]{@{}c@{}}{\bf \# of filters}\end{tabular}\\
 %Ethnicity & Male images & Female images \\
\hline
\hline
  conv0 & 3$\times$3/1/2 & 128 \\
  conv1 & 3$\times$3/1/2 & 1,024 \\
  RB1 & 3$\times$3/1/1 & 1,024\\
  fc1 & 512 & - \\
  fc2 & 16,384 & - \\
  \hline
  \hline
  conv2 & 3$\times$3/1/1 & 4*512\\
  PS1 & - & - \\
  conv3 & 5$\times$5/1/1 & 3\\
  \hline
\end{tabular}
\label{Tab:block8}
\end{small}
\end{center}
\end{table}

\begin{table}
\begin{center}
\captionsetup{justification=centering}
\caption{block\_16 architecture (input size is 16$\times$16$\times$3)}
\begin{small}
\begin{tabular}{  | c | c| c| }
\hline
\begin{tabular}[x]{@{}c@{}}{\bf Layer}\end{tabular} & \begin{tabular}[x]{@{}c@{}}{\bf Filter/Stride/Dilation}\end{tabular} & \begin{tabular}[x]{@{}c@{}}{\bf \# of filters}\end{tabular}\\
 %Ethnicity & Male images & Female images \\
\hline
\hline
  conv0 & 3$\times$3/1/2 & 128 \\
  conv1 & 3$\times$3/2/1 & 512\\
  RB1 & 3$\times$3/1/1 & 512\\
  conv2 & 3$\times$3/2/1 & 1,024 \\
  RB2 & 3$\times$3/1/1 & 1,024\\
  fc1 & 512 & - \\
  fc2 & 16,384 & - \\
  \hline
  \hline
  conv3 & 3$\times$3/1/1 & 4*512\\
  PS1 & - & - \\
  conv4 & 3$\times$3/1/1 & 4*256\\
  PS2 & - & - \\
  conv5 & 5$\times$5/1/1 & 3\\
  \hline
\end{tabular}
\label{Tab:block16}
\end{small}
\end{center}
\end{table}

\begin{table}
\begin{center}
\captionsetup{justification=centering}
\caption{block\_32 architecture (input size is 32$\times$32$\times$3)}
\begin{small}
\begin{tabular}{  | c | c| c| }
\hline
\begin{tabular}[x]{@{}c@{}}{\bf Layer}\end{tabular} & \begin{tabular}[x]{@{}c@{}}{\bf Filter/Stride/Dilation}\end{tabular} & \begin{tabular}[x]{@{}c@{}}{\bf \# of filters}\end{tabular}\\
 %Ethnicity & Male images & Female images \\
\hline
\hline
  conv0 & 3$\times$3/1/2 & 128 \\
  conv1 & 3$\times$3/2/1 & 256\\
  RB1 & 3$\times$3/1/1 & 256\\
  conv2 & 3$\times$3/2/1 & 512\\
  RB2 & 3$\times$3/1/1 & 512\\
  conv3 & 3$\times$3/2/1 & 1,024 \\
  RB3 & 3$\times$3/1/1 & 1,024\\
  fc1 & 512 & - \\
  fc2 & 16,384 & - \\
  \hline
  \hline
  conv3 & 3$\times$3/1/1 & 4*512\\
  PS1 & - & - \\
  conv4 & 3$\times$3/1/1 & 4*256\\
  PS2 & - & - \\
  conv5 & 3$\times$3/1/1 & 4*128\\
  PS3 & - & - \\
  conv6 & 5$\times$5/1/1 & 3\\
  \hline
\end{tabular}
\label{Tab:block32}
\end{small}
\end{center}
\end{table}

\begin{table}
\begin{center}
\captionsetup{justification=centering}
\caption{block\_64 architecture (input size is 64$\times$64$\times$3)}
\begin{small}
\begin{tabular}{  | c | c| c| }
\hline
\begin{tabular}[x]{@{}c@{}}{\bf Layer}\end{tabular} & \begin{tabular}[x]{@{}c@{}}{\bf Filter/Stride/Dilation}\end{tabular} & \begin{tabular}[x]{@{}c@{}}{\bf \# of filters}\end{tabular}\\
 %Ethnicity & Male images & Female images \\
\hline
\hline
  conv0 & 3$\times$3/1/2 & 128 \\
  conv1 & 3$\times$3/2/1 & 128\\
  RB1 & 3$\times$3/1/1 & 128\\
  conv2 & 3$\times$3/2/1 & 256\\
  RB2 & 3$\times$3/1/1 & 256\\
  conv3 & 3$\times$3/2/1 & 512\\
  RB3 & 3$\times$3/1/1 & 512\\
  conv4 & 3$\times$3/2/1 & 1,024 \\
  RB4 & 3$\times$3/1/1 & 1,024\\
  fc1 & 512 & - \\
  fc2 & 16,384 & - \\
  \hline
  \hline
  conv3 & 3$\times$3/1/1 & 4*512\\
  PS1 & - & - \\
  conv4 & 3$\times$3/1/1 & 4*256\\
  PS2 & - & - \\
  conv5 & 3$\times$3/1/1 & 4*128\\
  PS3 & - & - \\
  conv6 & 3$\times$3/1/1 & 4*64\\
  PS4 & - & - \\
  conv7 & 5$\times$5/1/1 & 3\\
  \hline
\end{tabular}
\label{Tab:block64}
\end{small}
\end{center}
\end{table}

\begin{table}
\begin{center}
\captionsetup{justification=centering}
\caption{block\_128 architecture (input size is 128$\times$128$\times$3)}
\begin{small}
\begin{tabular}{  | c | c| c| }
\hline
\begin{tabular}[x]{@{}c@{}}{\bf Layer}\end{tabular} & \begin{tabular}[x]{@{}c@{}}{\bf Filter/Stride/Dilation}\end{tabular} & \begin{tabular}[x]{@{}c@{}}{\bf \# of filters}\end{tabular}\\
 %Ethnicity & Male images & Female images \\
\hline
\hline
  conv0 & 3$\times$3/1/2 & 128 \\
  conv1 & 3$\times$3/2/1 & 64\\
  RB1 & 3$\times$3/1/1 & 64\\
  conv2 & 3$\times$3/2/1 & 128\\
  RB2 & 3$\times$3/1/1 & 128\\
  conv3 & 3$\times$3/2/1 & 256\\
  RB3 & 3$\times$3/1/1 & 256\\
  conv4 & 3$\times$3/2/1 & 512\\
  RB4 & 3$\times$3/1/1 & 512\\
  conv5 & 3$\times$3/2/1 & 1,024 \\
  RB5 & 3$\times$3/1/1 & 1,024\\
  fc1 & 512 & - \\
  fc2 & 16,384 & - \\
  \hline
  \hline
  conv3 & 3$\times$3/1/1 & 4*512\\
  PS1 & - & - \\
  conv4 & 3$\times$3/1/1 & 4*256\\
  PS2 & - & - \\
  conv5 & 3$\times$3/1/1 & 4*128\\
  PS3 & - & - \\
  conv6 & 3$\times$3/1/1 & 4*64\\
  PS4 & - & - \\
  conv7 & 3$\times$3/1/1 & 4*64\\
  PS5 & - & - \\
  conv8 & 5$\times$5/1/1 & 3\\
  \hline
\end{tabular}
\label{Tab:block128}
\end{small}
\end{center}
\end{table}

\begin{table*}
\begin{center}
\captionsetup{justification=centering}
\caption{Ablation Studies - quantitative results on the LFW \cite{LFW} dataset.}
\begin{small}
\begin{tabular}{  | c | c| c| c| c| }
\hline
\begin{tabular}[x]{@{}c@{}}{\bf Model}\end{tabular} & \begin{tabular}[x]{@{}c@{}}{\bf Mean Match Score}\end{tabular} & \begin{tabular}[x]{@{}c@{}}{\bf Mean SSIM} \cite{SSIM}\end{tabular} & \begin{tabular}[x]{@{}c@{}}{\bf FID} \cite{FID}\end{tabular} & \begin{tabular}[x]{@{}c@{}}{\bf Mean Perceptual Error} \cite{PieAPP}\end{tabular}\\
 %Ethnicity & Male images & Female images \\
\hline
\hline
  \begin{tabular}[x]{@{}c@{}}$l_2$ loss\end{tabular}  & 0.520 & 0.413 & 166.76 & 2.489 \\
  \hline
    \begin{tabular}[x]{@{}c@{}}w/o $L_{adv}$\end{tabular}  & 0.522 & 0.411 & 132.71 & 2.320 \\
    \hline
    \begin{tabular}[x]{@{}c@{}}w/o $L_{id}$\end{tabular}  & 0.609 & 0.519 & 91.65 & 1.956 \\
    \hline
    \begin{tabular}[x]{@{}c@{}}w/o $L_{pc}$\end{tabular}  & 0.624 & 0.528 & 101.44 & 2.046 \\
  \hline
    \begin{tabular}[x]{@{}c@{}}Ours (ProGAN)\end{tabular}  & 0.668 & 0.466 & 103.71 & 2.255 \\
    \hline
      \begin{tabular}[x]{@{}c@{}}Ours (Cascaded)\end{tabular}  & {\bf 0.722} & {\bf 0.753} & {\bf 46.12} & {\bf 1.256} \\
  \hline
\end{tabular}
\label{Tab:Ablation}
\end{small}
\end{center}
\vspace{-0.5cm}
\end{table*}

\section{Ablation Studies}
In this section, we analyze the effect of each component of our loss function on the overall quality of context and background synthesis. We present a comprehensive comparison that includes both qualitative results and quantitative experiments, using face images from the LFW dataset \cite{LFW}.

\begin{figure*}
\centering
  \includegraphics[width=1.0\linewidth]{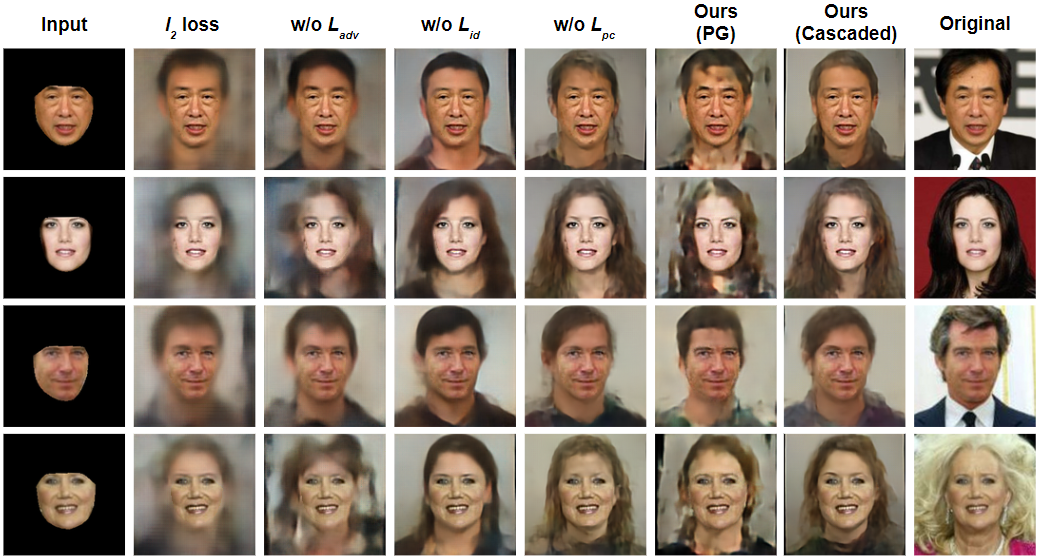}
  \caption{Ablation studies - hallucination results of our multi-scale GAN model and its variants.}
\label{fig:Ablation}
\end{figure*}

% \begin{table*}
% \begin{center}
% \captionsetup{justification=centering}
% \caption{Ablation studies - verification results on the LFW \cite{LFW} dataset.}
% \begin{small}
% \begin{tabular}{  | c | c| c| c| c| c| c| }
% \hline
% \begin{tabular}[x]{@{}c@{}}{\bf Model}\end{tabular} & \begin{tabular}[x]{@{}c@{}}{\bf $l_2$ loss}\end{tabular} & \begin{tabular}[x]{@{}c@{}}{\bf w/o $L_{adv}$}\end{tabular} &
% \begin{tabular}[x]{@{}c@{}}{\bf w/o $L_{id}$}\end{tabular} &
% \begin{tabular}[x]{@{}c@{}}{\bf w/o $L_{pc}$}\end{tabular} & \begin{tabular}[x]{@{}c@{}}{\bf Ours}\\{\bf (PG)}\end{tabular} & \begin{tabular}[x]{@{}c@{}}{\bf Ours}\\{\bf (Cascaded)}\end{tabular}\\
%  %Ethnicity & Male images & Female images \\
% \hline
%   \begin{tabular}[x]{@{}c@{}}TPR@\\FPR=0.01\end{tabular} & 0.732 & 0.747 & 0.816 & 0.808 & 0.811 & 0.842 \\
%   \hline
% \end{tabular}
% \label{Tab:Ablation}
% \end{small}
% \end{center}
% \vspace{-0.3cm}
% \end{table*}

For this experiment, we prepare four variations of our multi-scale cascaded GAN model, while keeping the network architecture intact. We replace $l_1$ loss with $l_2$ loss as the metric for computing $L_{pixel}$ for one model. For the other three models, we remove one of the other three losses (\ie, $L_{adv}$, $L_{id}$, and $L_{pc}$) in each case. We keep the weight of the other loss components intact in each case. To analyze the role of the training regime, we compare each of these cascaded models with our ProGAN model keeping other factors constant. For this experiment, we use the same set of quality metrics as before - (1) mean match score with ResNet-50 \cite{ResNet}, (2) mean SSIM \cite{SSIM}, (3) FID \cite{FID}, and (4) mean perceptual error \cite{PieAPP} (description of each metric is available in Section 4 of main text). The quantitative results are presented in Table \ref{Tab:Ablation}, along with visual results in Figure \ref{fig:Ablation}.

As expected, we find using $l_2$ loss for $L_{pixel}$ drastically deteriorates the quality of the hallucinated face images by producing blurrier results. Since the pixel intensities are normalized to [0, 1], $l_2$ loss suppresses high frequency signals, compared to $l_1$, due to its squaring operation. The absence of a discriminator (w/o $L_{adv}$) at a network block fails to push the results towards the distribution of real face images, consequently hampering the performance of the model. Although not as critical as $L_{pixel}$ and $L_{adv}$, the inclusion of both $L_{id}$ and $L_{pc}$ refine the hallucination result, as apparent from both the example images and the quality scores. The impact of the training regime, comparing end-to-end cascaded training with progressive growing (ProGAN), has already been discussed in Section 4 of the main text.

\section{Epoch by Epoch Learning}
\begin{figure}[t]
\centering
  \includegraphics[width=1.0\linewidth]{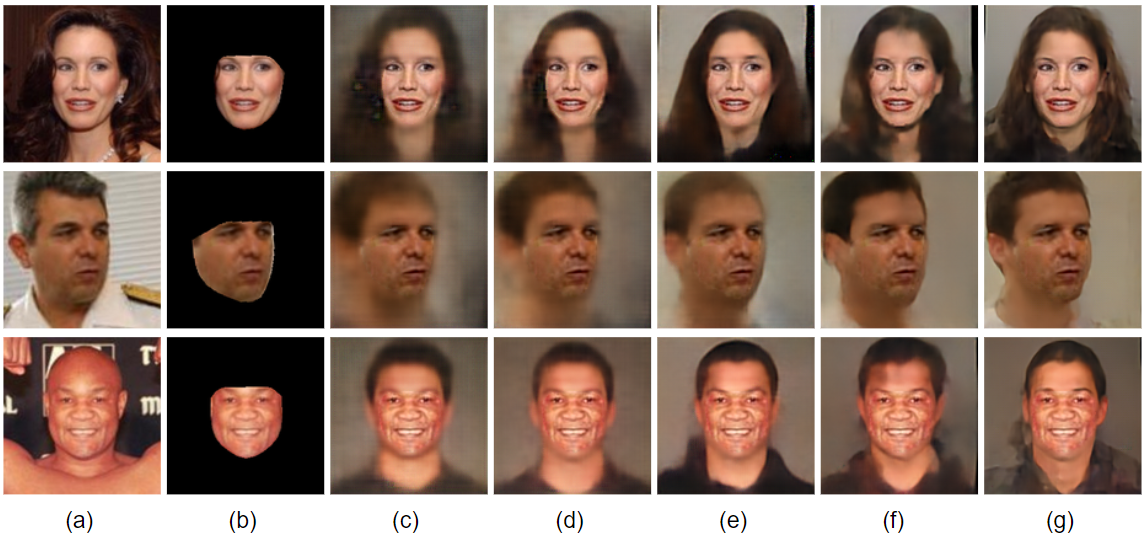}
  \caption{Sample synthesis results from LFW \cite{LFW} at different levels of training - (a) the original face image (cropped), (b) masked face input, hallucination results after (c) 10 epochs, (d) 20 epochs, (e) 30 epochs, (f) 40 epochs, and (g) 50 epochs of training.}
\label{fig:EpochByEpoch}
\vspace{-0.5cm}
\end{figure}

To understand how the context and background are learned by the model during training, we save snapshots of our cascaded GAN model at different levels of training - 10 epochs, 20 epochs, 30 epochs, 40 epochs and 50 epochs. Except the training iterations, all other parameters and hyper-parameters remain the same. These models are then used to generate context and background pixels on masked face images from LFW \cite{LFW}. Hallucinations for three such images have been shown in Figure \ref{fig:EpochByEpoch}.

As apparent from the figure, the model learns to generate a rough set of hair and skin pixels in the first few training epochs, not focusing on the clothes or background (10-20 epochs). Then it adds in pixels for the clothes and background, while further refining the overall skin and hair pixel quality (30-40 epochs). The validation loss stabilizes around the 50-th epoch (our hard termination point), and hence this snapshot has been used in our experiments. We also find the model to take a few extra iterations of refinement in hallucinating context and background for images with posed faces compared to those with frontal faces.

\section{Changing the Background Pixels}
To add more variety to our images, we add a post-processing step to further change the background pixels, while keeping the face and context pixels unchanged, using background images supplied by the user. We first locate the pixels outside the background (context + face mask) using the segmentation network from \cite{zhou2017scene,zhou2016semantic,xiao2018unified}. The pixels with the label 'Person' are kept inside the mask, which is further refined by a saliency map. This saliency map is computed using the gradient of each pixel of the image and the outer contour detected as the salient edge. The union of the initial mask and the points inside this contour produces the final foreground mask. Alternatively, the foreground mask can also be generated using the image matting network provided in \cite{Matting}. The new background image is then blended in with the help of this foreground mask using a Laplacian pyramid based blending \cite{LapPyr,SREFI1}.

\begin{figure}[t]
\centering
  \includegraphics[width=1.0\linewidth]{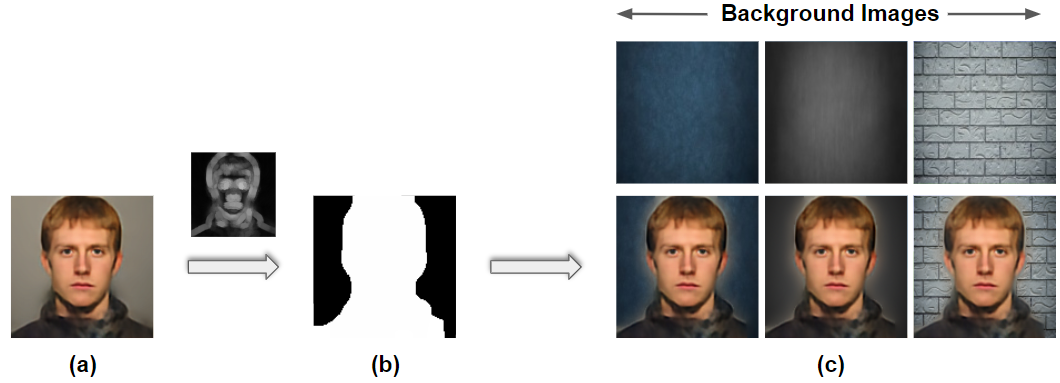}
  \caption{Background replacement process - (a) hallucinated face image (b) the detected foreground mask using a combination of gradient map and the segmentation network from \cite{zhou2017scene,zhou2016semantic,xiao2018unified}, and (c) background pixels replaced with Laplacian blending \cite{LapPyr}.}
\label{fig:ChangeBack}
\vspace{-0.6cm}
\end{figure}

\section{Additional Qualitative Results}
\begin{figure}[t]
\centering
  \includegraphics[width=1.0\linewidth]{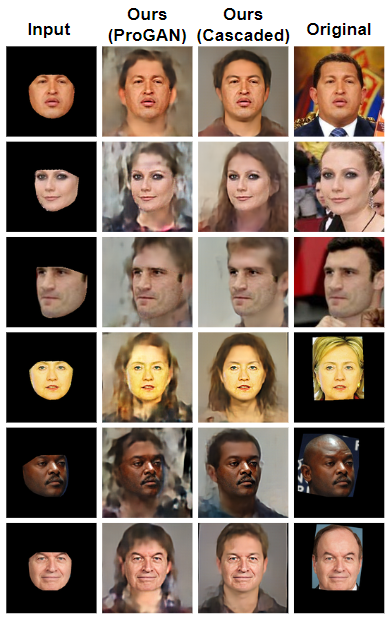}
  \caption{Additional qualitative results generated by our ProGAN and cascaded models. The first three rows are samples from the LFW \cite{LFW} dataset, while the last three rows are taken from the IJB-B \cite{IJBB} dataset. All images are 128$\times$128 in size.}
\label{fig:LFW_IJBB}
\vspace{-0.6cm}
\end{figure}

In this section, we present additional qualitative results for visual perusal. Face images, varying in gender, ethnicity, age, pose, lighting and expression, are randomly selected from the LFW dataset \cite{LFW} and IJB-B \cite{IJBB} video frames. Each image is then aligned about their eye centers using landmark points extracted from Dlib \cite{Dlib}, face masked and resized to 128$\times$128. Each image is then fed to the trained snapshots, used in our original experiments, of our cascaded and progressively growing models for context and background pixel synthesis. The results are shown in Figure \ref{fig:LFW_IJBB}.

% \begin{figure}[t]
% \centering
%   \includegraphics[width=1.0\linewidth]{Images/IJBB_Extra.png}
%   \caption{Additional qualitative results by the progressively growing (PG) and cascaded versions of our model on face images extracted from video frames from the IJB-B dataset \cite{IJBB}. All images are 128$\times$128 in size.}
% \label{fig:IJBB_Extra}
% \end{figure}

\section{Model Limitations}
\begin{figure}[t]
\centering
  \includegraphics[width=0.9\linewidth]{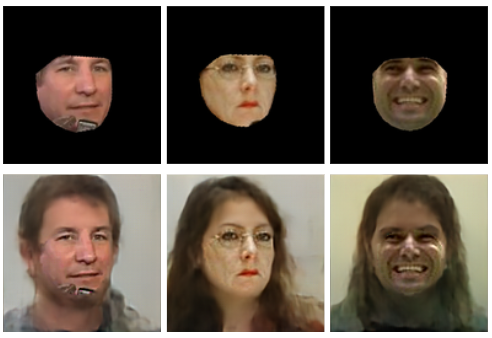}
  \caption{Some problematic cases - missing pixels for the microphone occluding subject's chin (left), no matching temples generated for the eyeglasses (middle), and hairstyle of wrong gender (right).}
\label{fig:Limits}
%\vspace{-0.58cm}
\end{figure}

As our model learns to hallucinate from the training data, we observe visual artifacts for face masks which vary drastically in appearance from it. For example, it fails to hallucinate missing pixels of occluding objects present in the face mask (like the microphone in leftmost image in Figure \ref{fig:Limits}). This can be fixed by refining the input face mask to remove such occluding objects. In some cases our model mis-labels the gender of the face mask and generates the wrong hairstyle. Such an example can be seen Figure \ref{fig:Limits} (rightmost image), where the input male subject gets a female hairstyle. This issue can be resolved by either training two networks separately with male and female subjects or by adding a gender preserving loss (using \cite{HassGen}) to the loss function. Our model also fails to generate matching temples when the subject wears eyeglasses due to their absence in the training images (Figure \ref{fig:Limits} middle image). To tackle this issue, the training data can be augmented by adding eyeglasses to some images using \cite{RSGAN,AttnGAN,StarGAN}.

% In some cases our model mis-labels the gender of the face mask and generates the wrong hairstyle. Such an example can be seen Figure \ref{fig:Limits} (rightmost image), where the input male subject gets a female hairstyle. This issue can be resolved by either training two networks separately with male and female subjects or by adding a gender preserving loss (using \cite{HassGen}) to the loss function.

\vspace{-0.3cm}
\section{Conclusion}
In this paper, we propose a cascaded network of GAN blocks that can synthesize realistic context and background pixels given a masked face input, without requiring any user supervision. Instead of swapping a source face onto a target image or inpainting small number of missing facial pixels, our model directly hallucinates the entire set of context and background pixels, by learning their representation directly from the training data. Each GAN block learns to hallucinate the missing pixels at a particular resolution via a combination of different losses and guides the synthesis process of the next block. 

While trained on only 12K face images acquired at a controlled setting, our model is effective in generating on challenging images from the LFW \cite{LFW} dataset. When compared with popular facial inpainting models \cite{GenFaceComp,DeepFill,symmFC,EdgeConn} and face swapping methods (DeepFake), our model generates more identity-preserving (evaluated using deep features from ResNet-50 \cite{ResNet}) and realistic (evaluated using SSIM \cite{SSIM}, FID \cite{FID}, and perceptual error \cite{PieAPP}) hallucinations. Our model can also be used to augment training data for CNNs by generating different hair and background of real subjects \cite{CASIA} or rendered synthetic face masks using \cite{MasiAug,SREFI2}. This can increase the intra-class variation in the training set, which in turn can make the CNN more robust to changes in hair and background along with variations in facial pose and shape. The generated face images can also be used as stock images by the media without any privacy concerns. 

A possible extension of this work would be to increase the resolution of the synthetic face images, possibly by adding more generator blocks to the cascaded network in a progressive manner \cite{ProgressiveGAN,HiResFaceComp}. The soft facial features of the generated output can also be varied by adding style based noise to the generator \cite{StyleGen}, while keeping the subject identity constant. Implementing this scheme to work on full face videos could be another avenue to explore. 

{\small
\bibliographystyle{ieee}
\bibliography{egbib}
}

\end{document}